\documentclass[twoside,leqno,twocolumn]{article}
\usepackage[letterpaper]{geometry} 
\usepackage{ltexpprt} 
\usepackage{graphicx} 
\usepackage{booktabs, multirow} 
\usepackage{mathtools, amssymb} 
\usepackage{algorithmic}
\usepackage{algorithm}
\usepackage{placeins}
\usepackage{xcolor}
\usepackage{subcaption}
\newcommand{\V}{${\mathcal V}$}
\newcommand{\E}{${\mathcal E}$}
\newcommand{\Z}{${\mathcal Z}$}
\newcommand{\Y}{${\mathcal Y}$}

\begin{document}

\title{Deperturbation of Online Social Networks \\
via Bayesian Label Transition
\thanks{Partially supported by NSF with grant number IIS-1909916.}
}
\author{Jun Zhuang 
\thanks{Indiana University-Purdue University Indianapolis, IN, USA. junz@iu.edu, alhasan@iupui.edu.} \\
\and Mohammad Al Hasan \footnotemark[2]
}
\date{}

\maketitle



\begin{abstract} \small\baselineskip=9pt
Online social networks (OSNs) classify users into different categories based on their online activities and interests, a task which is referred as a node classification task. Such a task can be solved effectively using Graph Convolutional Networks (GCNs). However, a small number of users, so-called perturbators, may perform random activities on an OSN, which significantly deteriorate the performance of a GCN-based node classification task.
Existing works in this direction defend GCNs either by adversarial training or by identifying the attacker nodes followed by their removal. However, both of these approaches require that the attack patterns or attacker nodes be identified first, which is difficult in the scenario when the number of perturbator nodes is very small.
In this work, we develop a GCN defense model, namely GraphLT 
\footnote{Our source code is publicly available on Github at \textbf{https://github.com/junzhuang-code/GraphLT}}
, which uses the concept of label transition. GraphLT assumes that perturbators' random activities deteriorate GCN's performance. To overcome this issue, GraphLT subsequently uses a novel Bayesian label transition model, which takes GCN's predicted labels and applies label transitions by Gibbs-sampling-based inference and thus repairs GCN's prediction to achieve better node classification. Extensive experiments on seven benchmark datasets show that GraphLT considerably enhances the performance of the node classifier in an unperturbed environment; furthermore, it validates that GraphLT can successfully repair a GCN-based node classifier with superior performance than several competing methods.
\end{abstract}

\section{Introduction}
\label{sec:intro}
In online social networks (OSNs), such as, Facebook and Twitter, a common exercise is user profiling, which clusters users into different groups based on their interests and online behavior. Such clusters are typically annotated so that a supervised node classification model can be built, which can subsequently be used for advertisement, product recommendation, and promotion offer generation. However, a small number of OSN users, so-called perturbators, often modify the network arbitrarily---for example, business users may randomly connect with many other users for commercial promotion. Activities of perturbators generally weaken the node classification model leading to poor performance. To improve the performance of the node classification under perturbations, such models should be defended so that the ramification of the perturbators' random activities can be mitigated.

Graph Convolutional Networks (GCNs) have been widely used on the node classification task \cite{bruna2013spectral, defferrard2016convolutional, kipf2016semi, wu2019simplifying, hamilton2017inductive} due to superior performance. However, GCNs are also shown to be vulnerable~\cite{wang2018attack, zugner2018adversarial} to adversarial attacks.
According to \cite{sun2018adversarial}, two kinds of defending methodologies have been proposed: the adversarial-based method \cite{jin2019latent, deng2019batch, feng2019graph, zhang2020defensevgae, xu2019topology,  entezari2020all, elinas2019variational}, and the detection-based method \cite{zhang2019comparing, wu2019adversarial, zheng2019robust}.
The former improves the robustness of GCNs by training with adversarial samples and the latter identifies the attacker nodes (or edges) and alleviates the negative impact by removing them. 
However, such approaches are not a good fit for defending GCNs from perturbators firstly because a small number of attack patterns are not easy to be discovered by OSN administrators, hence, adversarial samples cannot be used in the training phase. Also, perturbators are not Sybil users \cite{breuer2020friend}, i.e., they obey rules and don't conduct malicious attacks, and hence cannot be removed.

In this paper, we propose an alternative approach for defending GCNs from perturbators' actions. Instead of identifying perturbators and directly seeking the remedy of their actions, our proposed approach repairs the prediction of GCNs by assuming that the annotated node labels (which are observed and used for training the GCN) are noisy, which warrants the prediction of the GCN to be adjusted by using a methodological approach. 
Specifically, we propose a novel Bayesian label transition model, namely GraphLT (LT stands for Label Transition), that learns a label transition matrix, which enables the substitution of the predicted label of the GCN with an alternative label, if needed.
In the beginning, GraphLT trains the GCN-based node classifier with noisy labels on the train graph; the latent labels which would have been assigned to the nodes in the absence of perturbation, are unobserved. For each node, GCN returns a multinomial distribution over the label set. GraphLT then infers the label for each node by sampling from this multinomial distribution. The inference expects that inferred labels would match with the corresponding latent labels. In each iteration, the conditional label transition matrix is dynamically updated by replacing the predicted labels with the inferred labels from Gibbs sampling. With each subsequent iterations, the inferred labels increasingly align with the latent labels, making the inference more accurate. Note that, throughout the process, the latent labels are unknown, so GraphLT employs Bayesian inference to approximate the latent label distribution through conditional label transition. This is made possible by considering that the conditional label transition vector (a multinomial distribution) of each of the $K$ labels follows a Dirichlet prior, and these vectors are iteratively updated using the Bayesian framework.
The above design of GraphLT provides two main advantages. First, training a node classifier with noisy labels may cause the performance to decline, which GraphLT can reverse by applying appropriate label transition. Second, GraphLT can repair the prediction of the node classifier when the graph is perturbed with no malicious intent. Most importantly, GraphLT does not require identifying perturbator (or attacker), so it can be applied without any knowledge of the perturbator's activities or an attacker's attack model.
In the experiment section, we show results to demonstrate that GraphLT can improve the GCN-based node classifier's performance substantially in either of the scenarios, unperturbed graph and perturbed graph.
Overall, our contribution can be summarized as follows:
\begin{itemize}
  \item We propose a new Bayesian label transition model, namely GraphLT, to improve the performance of the node classifier on graph data. To the best of our knowledge, our work is the first model that adapts the Bayesian label transition method on GCNs for deperturbation in online social networks. 
  \item GraphLT can enhance the performance of node classification by Bayesian label transition under both unperturbed and perturbed environments. We also show that this enhancement is generalizable over three classic GCNs, and across various sizes of attributed graphs.
  \item Extensive experiments demonstrate that GraphLT is superior to the competing models on seven datasets of different sizes (two KDD Cup 2020 competition datasets and five public graph datasets).
\end{itemize}

\section{Methodology}
\label{sec:method}
In this section, we first introduce the notation, preliminary background of GCNs, and the basic idea of label transition. Furthermore, we theoretically analyze the Bayesian label transition. In the end, we explain the algorithm of our proposed model, GraphLT, and analyze its time complexity.

\noindent
\textbf{Notations and Preliminaries.}
In this paper, OSN is represented as an undirected attributed graph $\mathcal{G}$ = $(\mathcal{V}$, $\mathcal{E})$, where \V\ = $\{ v_{1}, v_{2}, ..., v_{N} \}$ denotes the set of vertices (users), $N$ is the number of vertices in $\mathcal{G}$, and \E\ $\subseteq$ \V\ $\times$ \V\ denotes the set of edges between vertices (connections among users). We denote $\mathbf{A} \in \mathbb{R}^{N \times N}$ as the symmetric adjacency matrix and $\mathbf{X} \in \mathbb{R}^{N \times d}$ as the feature matrix (contains users' profile information), where $d$ is the number of features for each vertex.
We assume that all ground-truth labels, hereby referred as {\bf latent labels} of the vertices $\mathcal{Z} \in \mathbb{R}^{N \times 1}$, are unobserved. We argue that manual annotation could be a potential solution to this problem but human annotation unavoidably brings into noises \cite{misra2016seeing}. We use $\mathcal{Y} \in \mathbb{R}^{N \times 1}$ to denote the manual-annotated {\bf noisy labels}, which are observed for all nodes (train and test). Our task is to defend GCN-based node classification when its noisy labels (observed) deviate from its latent labels (unobserved). However, we assume that the entries of both \Y\ and \Z\ take values from the same closed category set. Below, we first discuss the variant of graph convolutional networks (GCNs) that we consider for our task.
The most representative GCN proposed by Kipf and Welling \cite{kipf2016semi} is our preferred GCNs' variant. The layer-wise propagation of this GCN is presented as follows:
\begin{equation}
\footnotesize
\mathbf{H}^{(l+1)} = \sigma \left( \mathbf{\tilde{D}}^{-\frac{1}{2}} \mathbf{\tilde{A}} \mathbf{\tilde{D}}^{-\frac{1}{2}} \mathbf{H}^{(l)} \mathbf{W}^{(l)} \right)
\label{eqn:gcn}
\end{equation}
In Equation (\ref{eqn:gcn}), $\mathbf{\tilde{A}} = \mathbf{A} + I_{N}$, $\mathbf{\tilde{D}} = \mathbf{D} + I_{N}$, where $I_{N}$ is the identity matrix and $\mathbf{D}_{i,i} = \sum_{j} \mathbf{A}_{i,j}$ is the diagonal degree matrix. $\mathbf{H}^{(l)} \in \mathbb{R}^{N \times d}$ is the nodes hidden representation in the $l$-th layer, where $\mathbf{H}^{(0)} = \mathbf{X}$. $\mathbf{W}^{(l)}$ is the weight matrix in the $l$-th layer. $\sigma(\cdot)$ denotes a non-linear activation function, such as ReLU.

In the K-class node classification task, we denote $\mathbf{z}_{n}$ as the latent label of node $v_{n}$ and $\mathbf{y}_{n}$ as the corresponding noisy label ($\mathbf{y}_{nk}$ is a one-hot vector representation of $\mathbf{y}_n$). In our setting, the GCN can be trained by the following loss function $\mathcal{L}$:
\begin{equation}
\footnotesize
\mathcal{L} = - \frac{1}{N} \sum_{n=1}^{N} \mathcal{L}(\mathbf{y}_{n}, \textit{f}_{\theta}(v_{n})) = - \frac{1}{NK} \sum_{n=1}^{N} \sum_{k=1}^{K} \mathbf{y}_{nk} \ln \textit{f}_{\theta}(v_{nk})
\label{eqn:loss1}
\end{equation}
where $\textit{f}_{\theta}(\cdot) = softmax(\mathbf{H}^{(l)})$ is the prediction of node classifier parameterized by $\theta$.

Our idea is to improve the performance of the node classifier by Bayesian label transition, in which a transition matrix of size $K\times K$ is learned, which reflects a mapping from \Y\ to \Z. We represent such a matrix by $\phi$ (the same term is also used to denote a label to label mapping function). Under the presence of $\phi$, the loss function $\mathcal{L}$ in Equation (\ref{eqn:loss1}) can be rewritten as follows:
\begin{equation}
\footnotesize
\mathcal{L} = - \frac{1}{N} \sum_{n=1}^{N} \mathcal{L}(\mathbf{y}_{n}, \phi^{-1} \circ \textit{f}_{\theta}(v_{n}))
\label{eqn:loss2}
\end{equation}
However, learning accurate $\phi$ is a difficult task as the latent labels of the nodes are not available. So, in our method, the $\phi$ is iteratively updated to approximate the perfect one by using the Bayesian framework.

\begin{figure}[h]
  \centering
  \includegraphics[width=\linewidth]{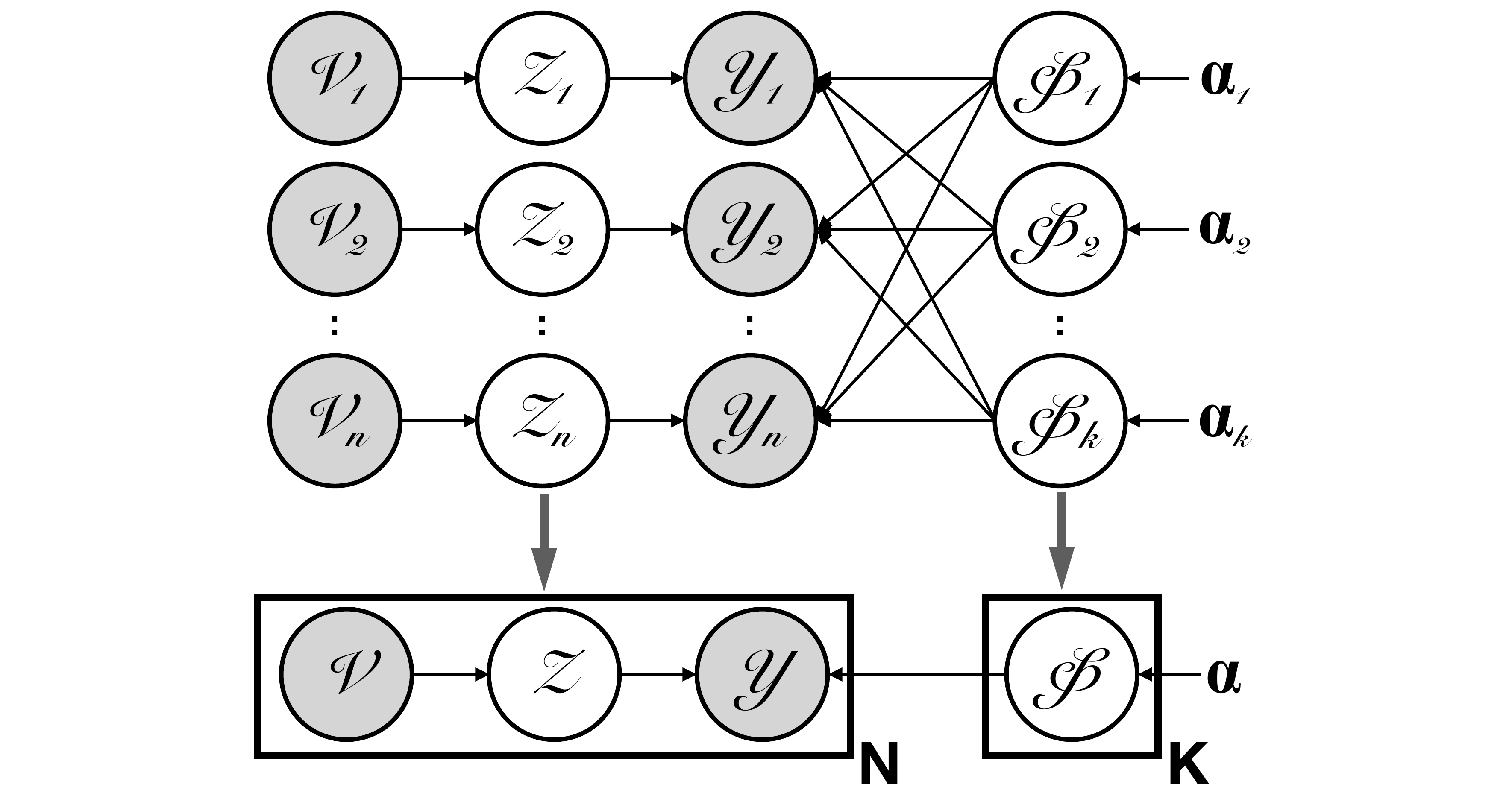}
  \caption{The Diagram of Bayesian Label Transition (\V, \Z\ and \Y\ denote the nodes, latent labels, and noisy labels, respectively. $\phi$ denotes the conditional label transition matrix. $\alpha$ denotes the Dirichlet parameter. $N$ and $K$ denote the number of nodes and classes, respectively.)}
\label{fig:fig_lt}
\end{figure}

\noindent
\textbf{Bayesian Label Transition.}
Figure \ref{fig:fig_lt} presents the diagram of Bayesian label transition. The unobserved latent labels (\Z) depend on the node features, whereas the observed noisy labels (\Y) depend on both \Z\ and the conditional label transition matrix, $\phi$, modeled by $K$ multinomial distributions, each with a Dirichlet prior having a parameter $\alpha_k$.
The latent label of node $v_n$, $\mathbf{z}_{n} \sim \textit{P} \left( \cdot \mid v_{n} \right)$, where $\textit{P} \left(\cdot \mid v_{n} \right)$ is a $Categorical$ distribution modeled by the node classifier $\textit{f}_{\theta}(v_{n})$.
The noisy label $\mathbf{y}_{n} \sim \textit{P} \left( \cdot \mid \phi_{\mathbf{z}_{n}} \right)$, where $\phi_{\mathbf{z}_{n}}$ is the parameter of $Categorical$ distribution $\textit{P} \left( \cdot \mid \phi_{\mathbf{z}_{n}} \right)$.
The conditional label transition matrix $\phi$ = $[\phi_{1}, \phi_{2}, …, \phi_{K}]^{T}$ $\in \mathbb{R}^{K \times K}$ consists of K transition vectors. The $k$-th transition vector $\phi_{k} \sim Dirichlet(\alpha_k)$, where $\alpha_k$ is the parameter of the $Dirichlet$ distribution associated with $k$-th transition vector. We use the symbol $\alpha$
to denote the set $\{\alpha_k\}_{1 \le k \le K}$. The goal of Bayesian label transition is to approximate the {\bf inferred label} of given node to the latent label of that node as identical as possible.

According to Figure \ref{fig:fig_lt}, we can employ Bayes' theorem to deduce the posterior of \Z\ which is conditioned on the node \V, the noisy labels \Y, and the Dirichlet parameter $\alpha$. We then continue to deduce the equations based on the conjugation property between the Multinomial distribution and the Dirichlet distribution as follows:
\begin{equation}
\scriptsize
\begin{aligned}
&\textit{P} \left( \mathcal{Z} \mid \mathcal{V}, \mathcal{Y} ; \alpha \right)
= \textit{P} \left( \phi ; \alpha \right) \cdot \textit{P} \left( \mathcal{Z} \mid \mathcal{V}, \mathcal{Y}, \phi \right) \\
&= \int_{\phi}   \prod_{k=1}^{K} \textit{P} \left( \phi_{k} ; \alpha_k \right) \cdot  \prod_{n=1}^{N} \textit{P} \left( \mathbf{z}_{n} \mid v_{n}, \mathbf{y}_{n}, \phi \right) d\phi \\
&= \int_{\phi}   \prod_{k=1}^{K} \textit{P} \left( \phi_{k} ; \alpha_k \right) \cdot   \prod_{n=1}^{N} \frac{\textit{P} \left( \mathbf{z}_{n} \mid v_{n} \right) \textit{P} \left( \mathbf{y}_{n} \mid \mathbf{z}_{n}, \phi \right)}{\textit{P} \left( \mathbf{y}_{n} \mid v_{n} \right)} d\phi \\
&= \prod_{n=1}^{N} \frac{\textit{P} \left( \mathbf{z}_{n} \mid v_{n} \right)}{\textit{P} \left( \mathbf{y}_{n} \mid v_{n} \right)} 
\int_{\phi}   \prod_{k=1}^{K} \frac{\Gamma \left( \sum_{k'=1}^{K} \alpha_{k'} \right)}{\prod_{k'=1}^{K} \Gamma \left( \alpha_{k'} \right)}   \prod_{k'=1}^{K} \phi_{kk'}^{\alpha_{k'}-1} \prod_{n=1}^{N} \phi_{\mathbf{z}_{n}\mathbf{y}_{n}}  d\phi \\
&= \prod_{n=1}^{N} \frac{\textit{P} \left( \mathbf{z}_{n} \mid v_{n} \right)}{\textit{P} \left( \mathbf{y}_{n} \mid v_{n} \right)}
\int_{\phi}   \prod_{k=1}^{K} \frac{\Gamma \left( \sum_{k'=1}^{K} \alpha_{k'} \right)}{\prod_{k'=1}^{K} \Gamma \left( \alpha_{k'} \right)}   \prod_{k'=1}^{K} \phi_{kk'}^{\mathbf{C}_{kk'} + \alpha_{k'} - 1} d\phi \\
&= \prod_{n=1}^{N} \frac{\textit{P} \left( \mathbf{z}_{n} \mid v_{n} \right)}{\textit{P} \left( \mathbf{y}_{n} \mid v_{n} \right)}
\prod_{k=1}^{K} \frac{\Gamma \left( \sum_{k'=1}^{K} \alpha_{k'} \right)}{\prod_{k'=1}^{K} \Gamma \left( \alpha_{k'} \right)}   \prod_{k=1}^{K} \frac{\prod_{k'=1}^{K} \Gamma \left( \alpha_{k'} + \mathbf{C}_{kk'}  \right)}{\Gamma \left( \sum_{k'=1}^{K} \left( \alpha_{k'} + \mathbf{C}_{kk'} \right) \right)} \\
\end{aligned}
\label{eqn:bayes}
\end{equation}
Here we denote the confusion matrix between the node prediction and the noisy labels as $\mathbf{C}$, where $\sum_{k}^{K} \sum_{k'}^{K} \mathbf{C}_{kk'} = N$. The term $ \prod_{n=1}^{N} \phi_{\mathbf{z}_{n}\mathbf{y}_{n}}$ is expressed as $ \prod_{k}^{K} \prod_{k'}^{K} \phi_{kk'}^{\mathbf{C}_{kk'}}$ so that we can integrate the terms based on the aforementioned conjugation property.

Unfortunately, Equation (\ref{eqn:bayes}) can not directly be employed to infer the label. Instead, we apply Gibbs sampling here to approximate our goal. According to Gibbs sampling, for each time we sample $\mathbf{z}_{n}$ by fixing $n$-th dimension in order to satisfy the detailed balance condition on the assumption of Markov chain. Combined with Equation (\ref{eqn:bayes}) and the recurrence relation of $\Gamma$ function, $\Gamma(n+1) = n\Gamma(n)$, we sample a sequence of $\mathbf{z}_{n}$ as follows:
\begin{equation}
\scriptsize
\begin{aligned}
\textit{P} \left( \mathbf{z}_{n} \mid \mathcal{Z}^{\neg \mathbf{z}_{n}}, \mathcal{V}, \mathcal{Y} ; \alpha \right) 
&= \frac{  \textit{P} \left( \mathcal{Z} \mid \mathcal{V}, \mathcal{Y} ; \alpha \right)  }{  \textit{P} \left( \mathcal{Z}^{\neg \mathbf{z}_{n}} \mid \mathcal{V}, \mathcal{Y} ; \alpha \right)  } \\
&= \frac{  \textit{P} \left( \mathbf{z}_{n} \mid v_{n} \right)  }{  \textit{P} \left( \mathbf{y}_{n} \mid v_{n} \right)  }
\frac{  \alpha_{\mathbf{y}_{n}} + \mathbf{C}_{\mathbf{z}_{n}\mathbf{y}_{n}}^{\neg \mathbf{z}_{n}}  }{  \sum_{k'=1}^{K} \left( \alpha_{k'} + \mathbf{C}_{\mathbf{z}_{n}k'}^{\neg \mathbf{z}_{n}} \right)  } \\
&\propto \overline{\textit{P}} \left( \mathbf{z}_{n} \mid v_{n} \right) \frac{  \alpha_{\mathbf{y}_{n}} + \mathbf{C}_{\mathbf{z}_{n}\mathbf{y}_{n}}^{\neg \mathbf{z}_{n}}  }{  \sum_{k'=1}^{K} \left( \alpha_{k'} + \mathbf{C}_{\mathbf{z}_{n}k'}^{\neg \mathbf{z}_{n}} \right) } \\
\end{aligned}
\label{eqn:gibbs}
\end{equation}
where we denote $\mathcal{Z}^{\neg \mathbf{z}_{n}}$ as the subset of $\mathcal{Z}$ that removes statistic $\mathbf{z}_{n}$. In the last row of Equation (\ref{eqn:gibbs}), the first term $\overline{\textit{P}} \left(\mathbf{z}_{n} \mid v_{n}\right)$ is a categorical distribution of labels for the node $v_n$ modeled by $f_\theta$. We use the term $\overline{\textit{P}} \left(\mathcal{Z} \mid \mathcal{V} \right) \in \mathbb{R}^{N \times K}$ to denote the same over all the nodes. Whereas the second term represents the conditional label transition which is obtained from the posterior of the multinomial distribution corresponding to label transition from $\mathbf{y}_n$ to $\mathbf{z}_n$. We use Equation (\ref{eqn:gibbs}) to sample the inferred label, $\mathbf{z}_n$. Also, $\phi$ is updated through Bayesian inference in each iteration. Such process is repeated for a given number of epochs with the expectation that subsequent inferred label can approximate to the latent label.

\begin{figure}[t]
  \centering
  \includegraphics[width=\linewidth]{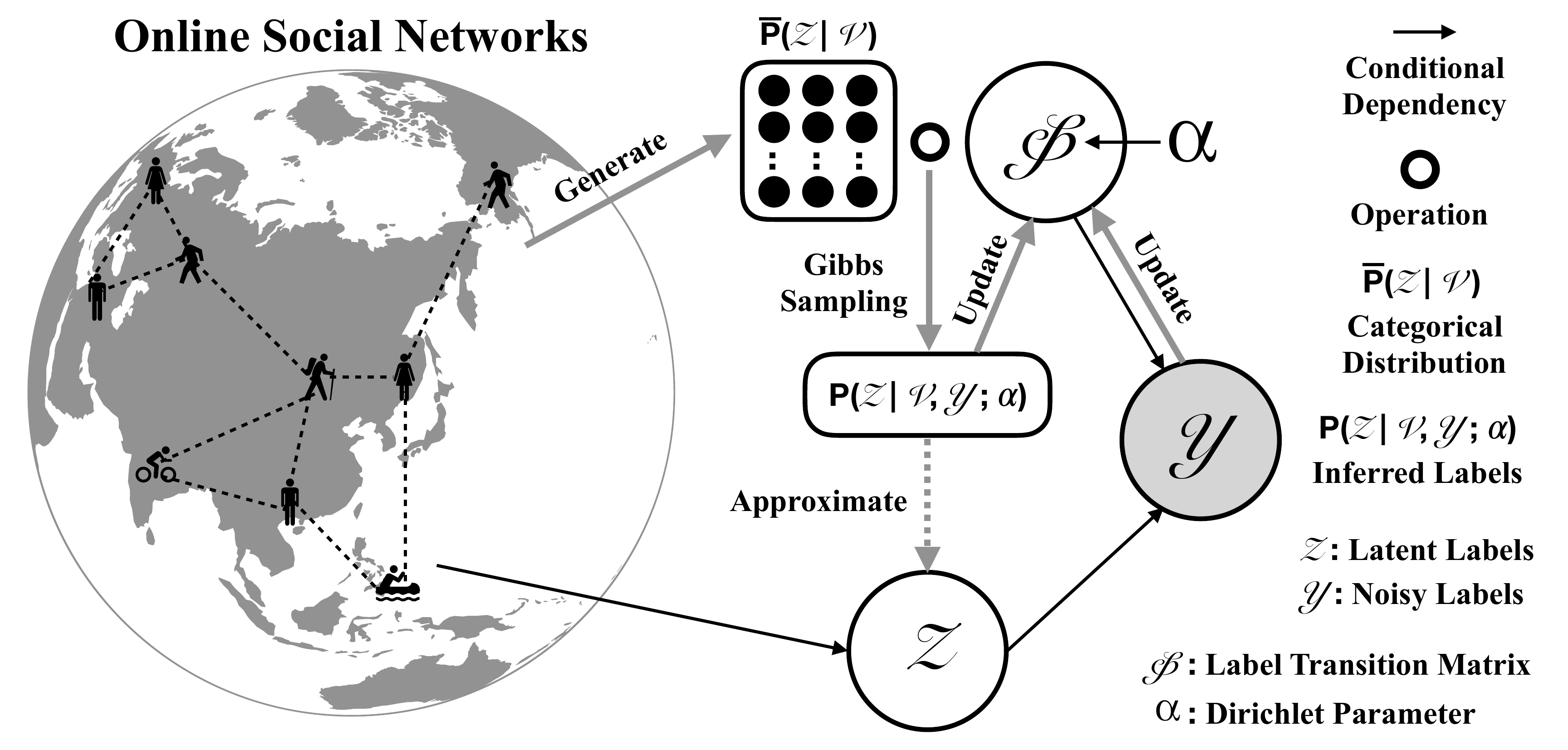}
  \caption{The Workflow of Bayesian Label Transition (GraphLT) (The left-hand side represents online social networks (OSN). Each human figure denotes a user of OSN. The dashed line between two users denotes the virtual connection in OSN. Note that the location of each user does not indicate the real location on earth. The latent labels represent the ground-truth labels of the nodes, which cannot be observed (White color). The noisy labels are manually annotated, which can be observed (Gray color).)}
\label{fig:model}
\end{figure}

\begin{algorithm}
\small
\caption{{\sc Bayesian} Label Transition}
\label{algo:graph_lt}
\hspace*{\algorithmicindent} \textbf{Input}: Graph $\mathcal{G}_{train}$ and $\mathcal{G}_{test}$, which contain corresponding symmetric adjacency matrix $\mathbf{A}$, feature matrix $\mathbf{X}$, and noisy labels \Y, Node classifier $\textit{f}_{\theta}$, The number of Warm-up steps $WS$, The number of inference epochs $Epochs$
\begin{algorithmic}[1]
\STATE{Train $\textit{f}_{\theta}$ by Equation (\ref{eqn:loss1}) on $\mathcal{G}_{train}$;}
\STATE{Generate categorical distribution $\overline{\textit{P}} \left(\mathcal{Z} \mid \mathcal{V} \right)$ by $\textit{f}_{\theta}$;}
\STATE{Compute warm-up label transition matrix $\phi'$ on $\mathcal{G}_{train}$;}
\STATE{Define the inferred labels $\textit{P} \left( \mathcal{Z} \mid \mathcal{V}, \mathcal{Y} ; \alpha \right) $ and dynamic label transition matrix $\phi$ on $\mathcal{G}_{test}$;}
\FOR{$step := 1$ $\textbf{to}$ $Epochs$}
    \IF{$step < WS$}
        \STATE{Sample $\mathbf{z}_{n}$ with warm-up $\phi'$ by Equation (\ref{eqn:gibbs});}
    \ELSE
        \STATE{Sample $\mathbf{z}_{n}$ with dynamic $\phi$ by Equation (\ref{eqn:gibbs});}
    \ENDIF
    \STATE{Update dynamic $\phi$ and $\textit{P} \left( \mathcal{Z} \mid \mathcal{V}, \mathcal{Y} ; \alpha \right) $;}
\ENDFOR
\RETURN{$\textit{P} \left( \mathcal{Z} \mid \mathcal{V}, \mathcal{Y} ; \alpha \right) $ and dynamic $\phi$;}
\end{algorithmic}
\end{algorithm}

\noindent
\textbf{GraphLT Algorithm and Pseudo-code.}
The total process of GraphLT is displayed in Figure \ref{fig:model}.
Assume that OSN presents an undirected attribute graph. GraphLT classifies the nodes and generates categorical distribution $\overline{\textit{P}} \left(\mathcal{Z} \mid \mathcal{V} \right) \in \mathbb{R}^{N \times K}$ at first. After that, GraphLT applies Gibbs sampling to sample the inferred labels $\textit{P} \left( \mathcal{Z} \mid \mathcal{V}, \mathcal{Y} ; \alpha \right) \in \mathbb{R}^{N \times 1}$ and updates the label transition matrix $\phi$ parameterized by $\alpha$. The information of \V\ is represented by both $\mathbf{A}$ and $\mathbf{X}$. The inference will ultimately converge, approximating the inferred labels $\textit{P} \left( \mathcal{Z} \mid \mathcal{V}, \mathcal{Y} ; \alpha \right) $ to the latent labels \Z\ as identical as possible. In brief, the goal of GraphLT is to sample the inferred labels by supervising the categorical distribution based on dynamic conditional label transition and ultimately approximates the inferred labels to the latent labels as identically as possible.

The pseudo-code of GraphLT is shown in Algorithm (\ref{algo:graph_lt}).
{\bf Training:} GraphLT trains the node classifier $\textit{f}_{\theta}$ on the train graph $\mathcal{G}_{train}$ at first (\textbf{Line 1}) and then generates categorical distribution $\overline{\textit{P}} \left(\mathcal{Z} \mid \mathcal{V} \right)$ by $\textit{f}_{\theta}$ (\textbf{Line 2}).
{\bf Inference:} Before the inference, GraphLT first computes a warm-up label transition matrix $\phi'$ by using the prediction over the train graph (\textbf{Line 3}) and then defines (creates empty spaces) the inferred labels $\textit{P} \left( \mathcal{Z} \mid \mathcal{V}, \mathcal{Y} ; \alpha \right) $ and the dynamic label transition matrix $\phi$ based on the test graph $\mathcal{G}_{test}$ (\textbf{Line 4}).
In the warm-up stage of the inference, GraphLT samples $\mathbf{z}_{n}$ with the warm-up label transition matrix $\phi'$ (\textbf{Line 7}), which is built with the categorical distribution of $f_\theta$ and the noisy labels on the train graph. The categorical distributions of both the train graph and the test graph should have high similarity if both follow a similar distribution. Thus, the warm-up $\phi'$ is a key-stone since subsequent inference largely depends on this distribution. 
After the warm-up stage, GraphLT samples $\mathbf{z}_{n}$ with the dynamic $\phi$ (\textbf{Line 9}). This dynamic $\phi$ updates in every epoch with current sampled $\mathbf{z}_{n}$ and corresponding $\mathbf{y}_{n} \in \mathcal{Y}$. Simultaneously, the inferred labels $\textit{P} \left( \mathcal{Z} \mid \mathcal{V}, \mathcal{Y} ; \alpha \right) $ is also updated based on the before-mentioned $\mathbf{z}_{n}$ (\textbf{Line 11}). The inference will ultimately converge, approximating the inferred labels to the latent labels as identically as possible.
Note that, both the train graph and the test graph contain corresponding symmetric adjacency matrix $\mathbf{A}$, feature matrix $\mathbf{X}$, and noisy labels \Y. The categorical distributions of the test graph may change abruptly when this graph is under perturbation since the original distribution in this graph is being perturbed. In this case, GraphLT can also help recover the original categorical distribution by dynamic conditional label transition.

According to Algorithm (\ref{algo:graph_lt}), GraphLT applies Gibbs sampling via Equation (\ref{eqn:gibbs}) inside the $FOR$ loop. The time complexity of the sampling is $\mathcal{O} ( N_{test} \times K + K^{2} )$ since element-wise multiplication only traverses the number of elements in matrices once, where $N_{test}$ denotes the number of nodes in the test graph. In practice, the number of test nodes is far more than the number of classes in OSN, i.e., $N_{test} \gg K$. So, the time complexity of this sampling operation is approximately equal to $\mathcal{O}(N_{test})$. Hence, the time complexity of inference except the training (\textbf{Line 1}) is $\mathcal{O}(Epochs \times N_{test})$, where $Epochs$ is the number of epochs for inference. This time complexity will be much lower than $\mathcal{O}(N_{test}^{2})$ when $Epochs$ is significantly smaller than $N_{test}$ in realistic scenario. This algorithm could be easily extended to the online inference version. We could reload the node classifier $\textit{f}_{\theta}$ that is pre-trained offline. To do so, the total time complexity approximates to $\mathcal{O}(Epochs \times N_{test})$.

\section{Experiments}
\label{sec:exp}
In this section, we present experimental results to examine GraphLT's performance. We first introduce experimental backgrounds and settings. We then investigate how node classification benefits from GraphLT. Besides, we validate the defending performance of GraphLT against competing defenders and visualize the results. Finally, we analyze model parameters, conduct an ablation study, and discuss limitations and future directions.

\begin{table}
\footnotesize
\centering
\setlength{\tabcolsep}{3pt}
  \caption{Statistics of Seven Benchmark Datasets ($\left| \mathcal{V} \right|$, $\left| \mathcal{E} \right|$, $\left| F \right|$, $\left| C \right|$, {\#Iter.} denotes the number of nodes, edges, features, classes, training epochs, respectively. {Avg.D} denotes the average degree of test nodes.)}
  \label{tab:dataset}
  \begin{tabular}{ccccccc}
    \toprule
    \textbf{Dataset} & {$\left| \mathcal{V} \right|$} & {$\left| \mathcal{E} \right|$} & {$\left| F \right|$} & {$\left| C \right|$} & {\#Iter.} & {Avg.D} \\
    \midrule
    \textbf{KDD20(S1)} & 593,486 & 6,217,004 & 100 & 18 & 1,000 & 9.63 \\
    \textbf{KDD20(S2)} & 659,574 & 5,757,154 & 100 & 18 & 1,000 & 8.07 \\
    \textbf{Cora} & 2,708 & 10,556 & 1,433 & 7 & 200 & 3.85 \\
    \textbf{Citeseer} & 3,327 & 9,228 & 3,703 & 6 & 200 & 2.78 \\
    \textbf{AMZcobuy} & 13,752 & 574,418 & 767 & 10 & 500 & 37.07 \\
    \textbf{Coauthor} & 18,333 & 327,576 & 6,805 & 15 & 500 & 10.01 \\
    \textbf{Reddit} & 232,965 & 114,615,892 & 602 & 41 & 500 & 491.88 \\
  \bottomrule
\end{tabular}
\end{table}

\begin{table*}[h] 
\footnotesize
\centering
\setlength{\tabcolsep}{2.8pt}
\caption{The Investigation of How GraphLT Reverses The Performance’s Decline Caused by Training Node Classifiers $\textit{f}_{\theta}$ with Noisy Labels under Different Noise Ratio $nr$ ($\textbf{Orig.}$ denotes the original accuracy. $\textbf{LT}$ denotes the accuracy after label transition.)}
\label{table:exp2}
\begin{tabular}{c|c|cc|cc|cc|cc|cc|cc|cc} 
\toprule 
\multirow{2}{*}{} & \multirow{2}{*}{\parbox{1.3cm}{\centering \textbf{Test Acc (\%)}}} & 
  \multicolumn{2}{c|}{\textbf{KDD20(S1)}} &
  \multicolumn{2}{c|}{\textbf{KDD20(S2)}} &
  \multicolumn{2}{c|}{\textbf{Cora}} &
  \multicolumn{2}{c|}{\textbf{Citeseer}} &
  \multicolumn{2}{c|}{\textbf{AMZcobuy}} &
  \multicolumn{2}{c|}{\textbf{Coauthor}} &
  \multicolumn{2}{c}{\textbf{Reddit}} \\
\cline{3-16} 
  &  & \textbf{Orig.} & \textbf{LT} & \textbf{Orig.} & \textbf{LT} & \textbf{Orig.} & \textbf{LT} & \textbf{Orig.} & \textbf{LT} & \textbf{Orig.} & \textbf{LT} & \textbf{Orig.} & \textbf{LT} & \textbf{Orig.} & \textbf{LT}\\
\midrule

\multirow{4}{*}{\textbf{GCN}} & $nr$ = 0.3 & 35.04 & 70.48 & 60.93 & 76.37 & 81.30 & 77.74 & 66.47 & 76.78 & 89.48 & 75.11 & 91.96 & 76.75 & 92.91 & 85.15 \\
& $nr$ = 0.2 & 35.19 & 79.36 & 61.02 & 85.24 & 82.78 & 88.81 & 67.97 & 85.99 & 89.63 & 85.70 & 92.07 & 87.84 & 93.53 & 92.23 \\
& $nr$ = 0.1 & 35.92 & 88.79 & 62.82 & 91.49 & 83.03 & 94.22 & 69.57 & 93.19 & 89.58 & 93.75 & 92.29 & 95.51 & 93.61 & 96.61 \\
& $nr$ = 0.0 & 36.48 & 96.52 & 63.19 & 98.20 & 86.23 & 97.90 & 69.77 & 93.89 & \textbf{89.85} & \textbf{98.54} & 92.74 & 98.50 & 94.34 & 98.24 \\
\midrule

\multirow{3}{*}{\textbf{SGC}} & $nr$ = 0.3 & 35.83 & 64.98 & 59.58 & 73.22 & 83.27 & 79.46 & 68.07 & 76.77 & 82.31 & 75.64 & 90.51 & 76.73 & 91.41 & 85.01 \\
& $nr$ = 0.2 & 36.11 & 76.34 & 59.90 & 85.01 & 84.38 & 87.21 & 70.17 & 85.98 & 83.19 & 82.62 & 90.87 & 87.73 & 92.23 & 92.25 \\
& $nr$ = 0.1 & 36.21 & 86.47 & 60.11 & 90.15 & 84.87 & 95.20 & 70.47 & 93.59 & 84.25 & 92.85 & 91.36 & 95.46 & 92.44 & 96.71 \\
& $nr$ = 0.0 & 36.92 & 96.63 & 60.30 & 97.66 & 85.73 & 97.78 & 71.27 & 96.88 & 85.44 & 97.86 & 91.94 & 98.42 & 93.25 & 97.90 \\
\midrule

\multirow{3}{*}{\textbf{GraphSAGE}} & $nr$ = 0.3 & 57.09 & 79.20 & 62.19 & 79.68 & 82.41 & 78.11 & 69.67 & 76.68 & 85.96 & 77.12 & 91.27 & 79.53 & 94.01 & 85.16 \\
& $nr$ = 0.2 & 57.76 & 82.34 & 62.67 & 86.51 & 84.50 & 87.95 & 71.57 & 87.29 & 87.20 & 86.89 & 91.78 & 90.38 & 95.02 & 91.89 \\
& $nr$ = 0.1 & 57.89 & 90.13 & 63.01 & 91.52 & 85.98 & 96.19 & 72.47 & 93.69 & 87.83 & 94.21 & 93.02 & 96.66 & 95.21 & 96.23 \\
& $nr$ = 0.0 & \textbf{58.20} & \textbf{98.69} & \textbf{63.24} & \textbf{98.32} & \textbf{86.47} & \textbf{99.26} & \textbf{73.88} & \textbf{98.29} & 89.79 & 98.42 & \textbf{94.07} & \textbf{99.29} & \textbf{96.14} & \textbf{99.38} \\
\bottomrule
    \end{tabular}
\end{table*}

\begin{table*}[t] 
\footnotesize
\centering
\setlength{\tabcolsep}{3pt}
\caption{The Examination of The Generalization of GraphLT over Three Classic Node Classifiers ($nr$=0.1)}
\label{table:exp1}
\begin{tabular}{c|c|c|c|c|c|c|c|c} 
\toprule 
\textbf{} & \textbf{Test Acc (\%)} & \textbf{KDD20(S1)} & \textbf{KDD20(S2)} & \textbf{Cora} & \textbf{Citeseer} & \textbf{AMZcobuy} & \textbf{Coauthor} & \textbf{Reddit} \\
\midrule
\multirow{3}{*}{\textbf{GCN}} & \textcolor{blue}{\bf Before Perturbation} & 35.92 & 62.82 & 83.03 & 69.57 & \textcolor{blue}{\bf 89.58} & 92.29 & 93.61 \\
& \textcolor{gray}{\bf After Perturbation} & 28.89 & 33.54 & 27.80 & 19.02 & \textcolor{gray}{\bf 85.14} & \textcolor{gray}{\bf 35.75} & 78.45 \\ 
& \textcolor{red}{\bf After Label Transition} & 72.46 & 66.02 & 27.81 & 19.92 & \textcolor{red}{\bf 92.51} & 48.62 & 87.31 \\
\midrule
\multirow{3}{*}{\textbf{SGC}} & \textcolor{blue}{\bf Before Perturbation} & 36.21 & 60.11 & 84.87 & 70.47 & 84.25 & 91.36 & 92.44 \\
& \textcolor{gray}{\bf After Perturbation} & 25.56 & 28.68 & 27.91 & \textcolor{gray}{\bf 30.83} & 74.04 & 35.64 & 39.82 \\
& \textcolor{red}{\bf After Label Transition} & 56.43 & 57.93 & 27.93 & \textcolor{red}{\bf 37.84} & 90.21 & \textcolor{red}{\bf 59.66} & 62.37 \\
\midrule
\multirow{3}{*}{\textbf{GraphSAGE}} & \textcolor{blue}{\bf Before Perturbation} & \textcolor{blue}{\bf 57.89} & \textcolor{blue}{\bf 63.01} & \textcolor{blue}{\bf 85.98} & \textcolor{blue}{\bf 72.47} & 87.83 & \textcolor{blue}{\bf 93.02} & \textcolor{blue}{\bf 95.21} \\
& \textcolor{gray}{\bf After Perturbation} & \textcolor{gray}{\bf 50.23} & \textcolor{gray}{\bf 39.21} & \textcolor{gray}{\bf 27.92} & 19.82 & 50.95 & 21.60 & \textcolor{gray}{\bf 87.23} \\
& \textcolor{red}{\bf After Label Transition} & \textcolor{red}{\bf 76.34} & \textcolor{red}{\bf 74.26} & \textcolor{red}{\bf 27.95} & 20.32 & 60.64 & 24.64 & \textcolor{red}{\bf 95.87} \\
\bottomrule
    \end{tabular}
\end{table*}

\noindent
\textbf{Datasets.}
Table \ref{tab:dataset} shows statistics of seven node-labeled graphs. Both \textbf{KDD20(S1)} and \textbf{KDD20(S2)} are obtained from stage one (S1) and stage two (S2) of KDD Cup 2020 regular machine learning competition track 2, Adversarial Attacks and Defense on Academic Graph (KDD20 Competition) \footnote{https://www.kdd.org/kdd2020/kdd-cup}.
\textbf{Reddit} dataset is constructed by connecting Reddit posts if the same user comments on both posts \cite{hamilton2017inductive}.
\textbf{Coauthor} is co-authorship graphs of computer science based on the Microsoft Academic Graph \footnote{https://www.kdd.org/kdd-cup/view/kdd-cup-2016}.
\textbf{AMZcobuy} comes from the computer segment of the Amazon co-purchase graph \cite{shchur2018pitfalls}.
Both \textbf{Cora} and \textbf{Citeseer} are well-known citation graph data \cite{sen2008collective}.
Among the above datasets, KDD20(S1), KDD20(S1), and Reddit are large graphs. AMZcobuy and Coauthor are medium graphs. Cora and Citeseer are small graphs.

\noindent
\textbf{Competing Defending Models.}
We compare our model with state-of-the-art defending methods which work on graph convolutional networks to overcome graph perturbations.
{\bf RGCN}~\cite{zhu2019robust} adopts Gaussian distributions as the hidden representations of nodes to mitigate the negative effects of adversarial attacks.
{\bf GRAND}~\cite{feng2020graph} proposes random propagation and consistency regularization strategies to address the issues about over-smoothing and non-robustness of GCNs.
{\bf ProGNN}~\cite{jin2020graph} jointly learns the structural graph properties and iteratively reconstructs the clean graph to reduce the effects of adversarial structure. 
{\bf GNNGUARD}~\cite{zhang2020gnnguard} employs neighbor importance estimation and the layer-wise graph memory to quantify the message passing between nodes to defend the GCNs from adversarial attacks.
Further implementation details can be found in supplementary materials.

\noindent
\textbf{Experimental Settings.}
For all seven graphs, the nodes are partitioned into train, validation, and test, each comprising 40\%, 30\%, and 30\% of the nodes, respectively. We select such partitions to simulate the OSNs as the size of the existing users in OSNs will not be too small compared to the new coming users.
Since GraphLT works with noisy labels, for a subset of nodes, we generate noisy labels by randomly replacing the ground-truth label of a node with another label, chosen uniformly. We denote the percentage of such replacement as noise ratio $nr$.
We examine the generalization of GraphLT over different variants of graph convolutional networks, spectral methods (GCN \cite{kipf2016semi}, SGC \cite{wu2019simplifying}) and spatial methods (GraphSAGE \cite{hamilton2017inductive}).
To test the defending performance, we simulate non-malicious perturbations in OSN. We assume that perturbators intend to randomly connect with many other users for commercial promotion, which is a common behavior in realistic scenarios. Note that, the perturbations apply to the validation/test graphs only. To ensure that the perturbation is unnoticeable, we limit the number of perturbators to 1\% of the validation/test nodes (a.k.a. victim nodes). For each perturbator, we also limit the number of connections up to a given budget, which is 100. This simulation is similar to \cite{wang2018attack}. However, our simulation does not apply gradient-based attack, such as FGSM \cite{goodfellow2014explaining}, PGD \cite{madry2017towards}, etc.
Note that, the goal of this simulation is not to maliciously attack the node classifier. Instead, we want to examine the defending performance of GraphLT under this non-malicious perturbation setting.

\begin{table*}[t] 
\footnotesize
\centering
\setlength{\tabcolsep}{8pt}
\caption{The Comparison of Defending Performance between GraphLT ($nr$=0.1) and Competing Defenders}
\label{table:performance}
\begin{tabular}{c|c|c|c|c|c|c|c} 
\toprule 
\textbf{} & \textbf{KDD20(S1)} & \textbf{KDD20(S2)} & \textbf{Cora} & \textbf{Citeseer} & \textbf{AMZcobuy} & \textbf{Coauthor} & \textbf{Reddit} \\
\midrule
\textbf{RGCN}~\cite{zhu2019robust} & 70.97 & 69.62 & 26.12 & 20.09 & 36.16 & 23.96 & 79.30 \\
\textbf{GRAND}~\cite{feng2020graph} & 68.51 & 66.84 & 27.92 & \textcolor{red}{\bf 21.42} & 53.67 & 24.38 & 86.31 \\
\textbf{ProGNN}~\cite{jin2020graph} & 69.34 & 67.80 & 25.96 & 19.74 & 56.39 & 24.17 & 85.25 \\
\textbf{GNNGUARD}~\cite{zhang2020gnnguard} & 71.73 & 70.14 & 24.77 & 20.12 & 57.64 & 23.91 & 86.83 \\
\textbf{GraphLT} & \textcolor{red}{\bf 76.34} & \textcolor{red}{\bf 74.26} & \textcolor{red}{\bf 27.95} & 20.32 & \textcolor{red}{\bf 60.64} & \textcolor{red}{\bf 24.64} & \textcolor{red}{\bf 95.87} \\
\bottomrule
    \end{tabular}
\end{table*}

\noindent
\textbf{Node Classification Benefits from GraphLT.}
We first investigate how GraphLT reverses the performance's decline caused by training a node classifier with noisy labels. We examine our model with three node classifiers under different noise ratios, where $nr = [0.0, 0.1, 0.2, 0.3]$. Note that we don't apply perturbations in this examination. As shown in Table \ref{table:exp2}, we highlight the best performance for two scenarios, original performance (Orig.) and the performance after label transition (LT) on each dataset. It's expected that the accuracy increases as the noise ratio decreases. 
We observe that label transition still works well in some cases even if we use noisy labels with a higher noise ratio. For instance, we train GraphSAGE using the noisy labels with $nr=0.3$ and get 69.67\% original accuracy on Citeseer. This accuracy increases to 76.68\% after label transition (under the same noise ratio). This property is very helpful in real life. For example, the profile of new users may lack sufficient information and thus leads to inaccurate annotation. This property indicates that our model could still help improve the performance of the node classifier well even if we only acquire the annotated labels with a higher noise ratio.
We also observe that GraphSAGE cannot always outperform GCN. This phenomenon shows more significance on denser graphs. For example, GCN achieves the best classification performance on AMZcobuy. GCN also surpasses GraphSAGE after label transition ($nr$ = 0.2 \& 0.1) on Reddit. We argue that this phenomenon may be related to the degree of nodes. Compared to GCN, which applies 1-order polynomial on layer-wise graph convolution, GraphSAGE aggregates features from fixed-size local neighbors of the current node. Such aggregation may lose more information when the degree of nodes increases. That is to say, GCN may outperform GraphSAGE when the graph is denser (higher degrees of nodes).
We also examine the generalization of our proposed model over various variants of graph convolutional networks, specifically, GCN \cite{kipf2016semi}, SGC \cite{wu2019simplifying} and GraphSAGE \cite{hamilton2017inductive}, when the graph is under perturbed. As presented in Table \ref{table:exp1}, we compare the test accuracy of these node classifiers over three scenarios: Before perturbation, After perturbation, and After label transition. For all datasets, the performance of all node classifiers drops after perturbation. However, they improve substantially after we apply GraphLT's label transition. On some occasions, the accuracy after label transition is even better than the accuracy before perturbation on large graphs, whereas this improvement is not so obvious on small graphs. We argue that this phenomenon is caused by the size of the initial warm-up label transition matrix $\phi'$. The model can obtain a stronger $\phi'$ on a large graph so that its improvement gets better. Overall, all three node classifiers benefit from our proposed model, GraphLT. The performance of GraphLT is associated with the noise ratio and the size of the graph.

\begin{figure}[t] 
  \centering
  \includegraphics[width=\linewidth]{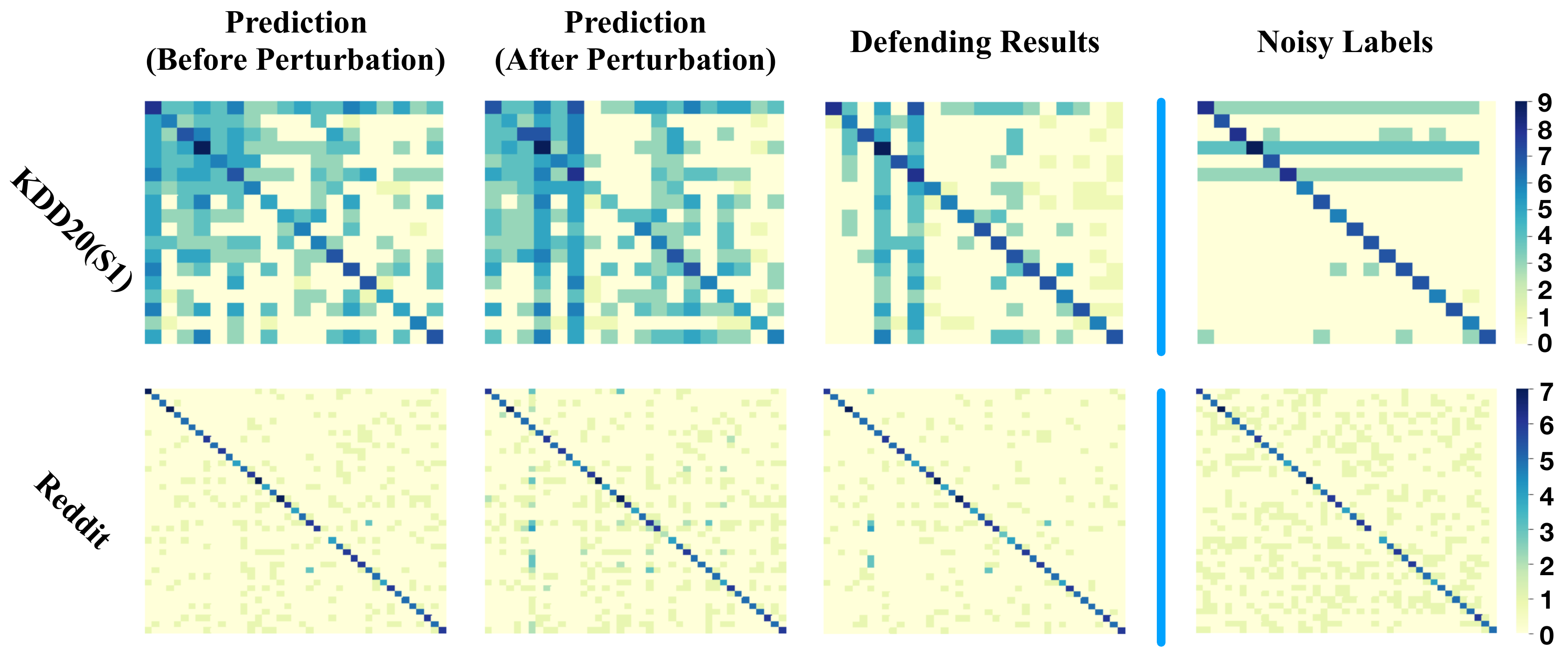}
  \caption{The Confusion Matrices (Heatmap) of The Defending Results on KDD20(S1) and Reddit for GraphLT ($nr$=0.1) (We apply log-scale to the confusion matrix for fine-grained visualization.)}
\label{fig:fig_cm}
\end{figure}

\begin{figure*}[t] 
  \begin{subfigure}{0.14\textwidth}
  \centering 
    \includegraphics[width=\linewidth]{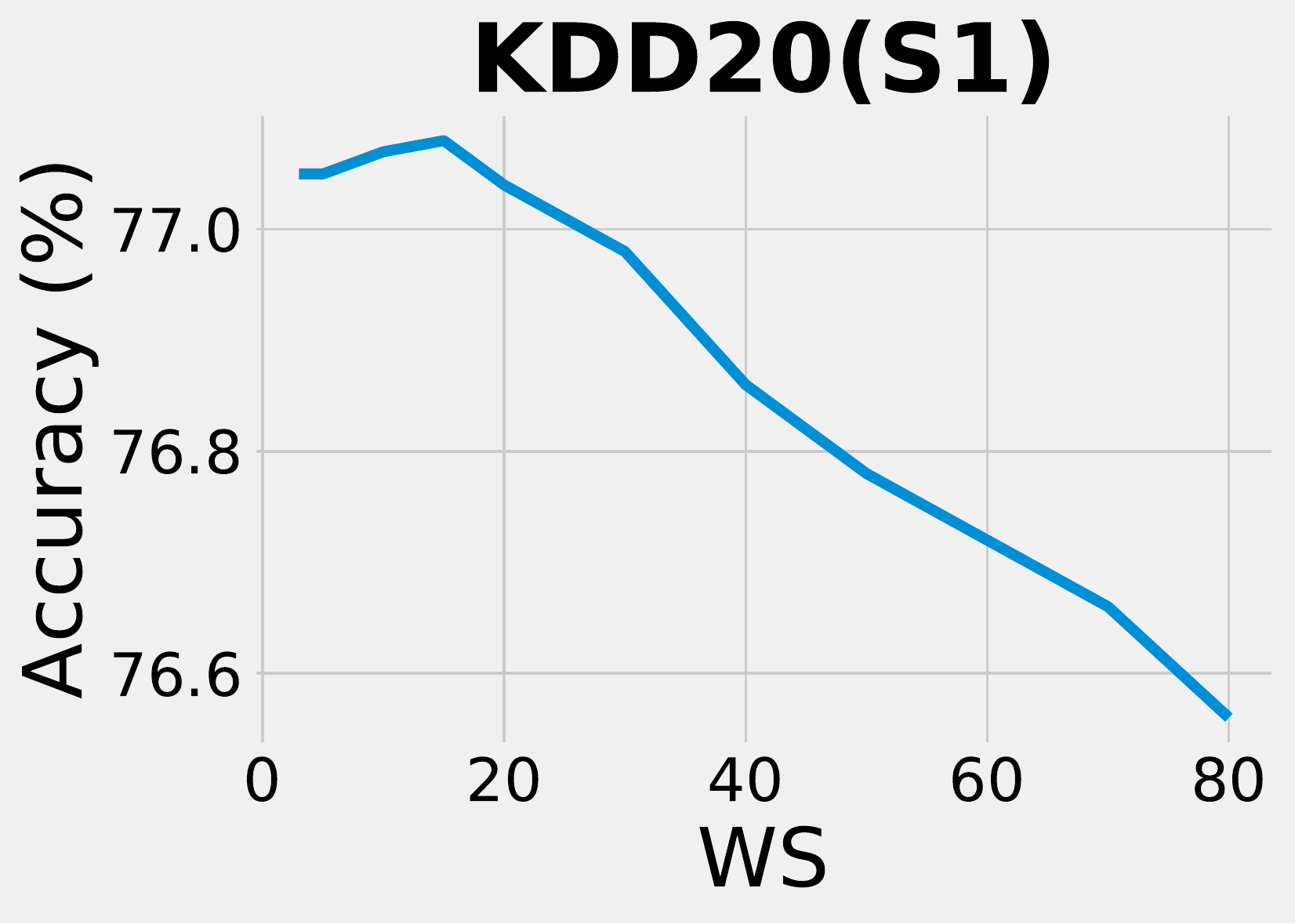}
  \end{subfigure}%
  \hfill
  \begin{subfigure}{0.14\textwidth}
  \centering 
    \includegraphics[width=\linewidth]{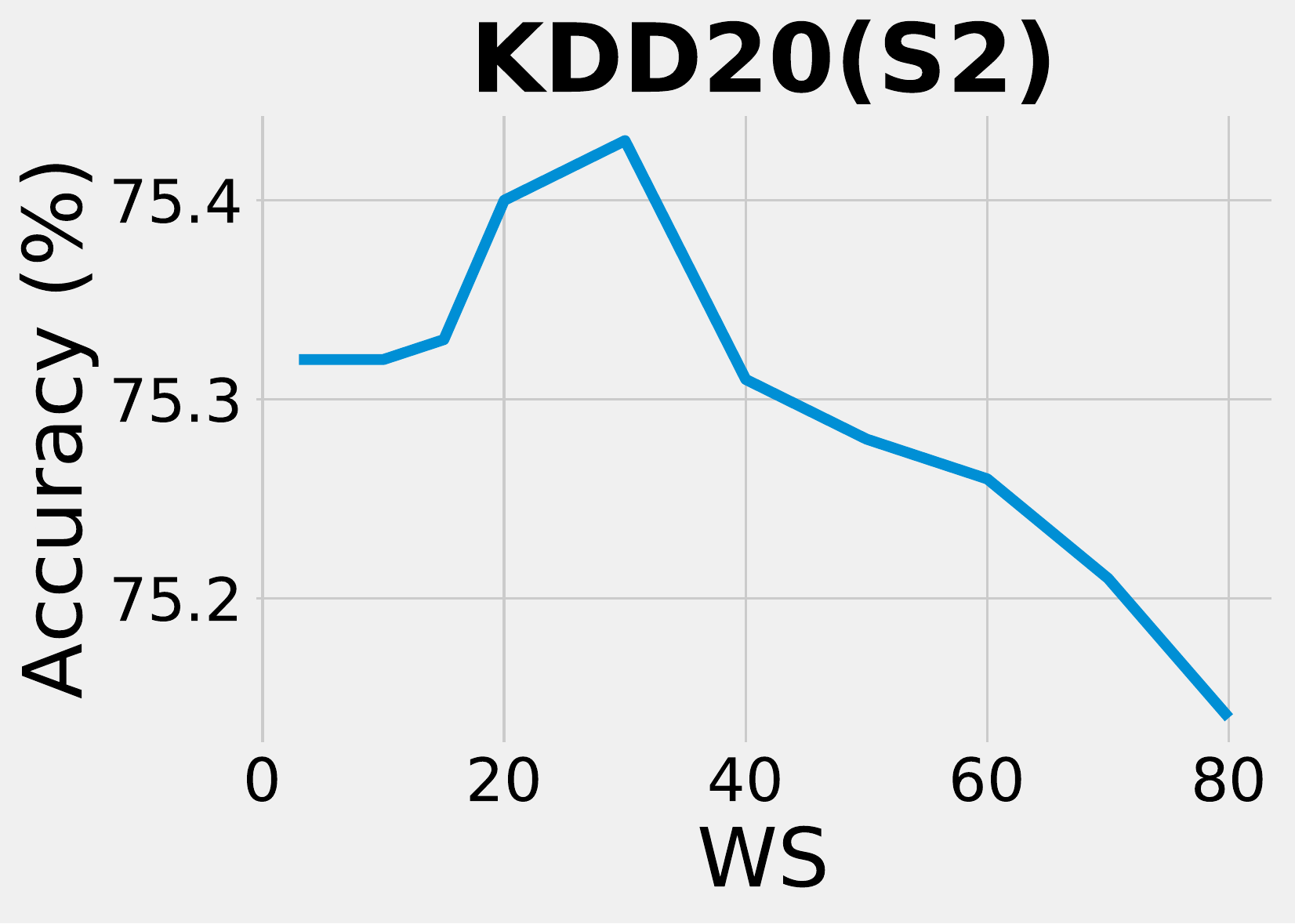}
  \end{subfigure}%
  \hfill
  \begin{subfigure}{0.1346\textwidth} 
  \centering 
    \includegraphics[width=\linewidth]{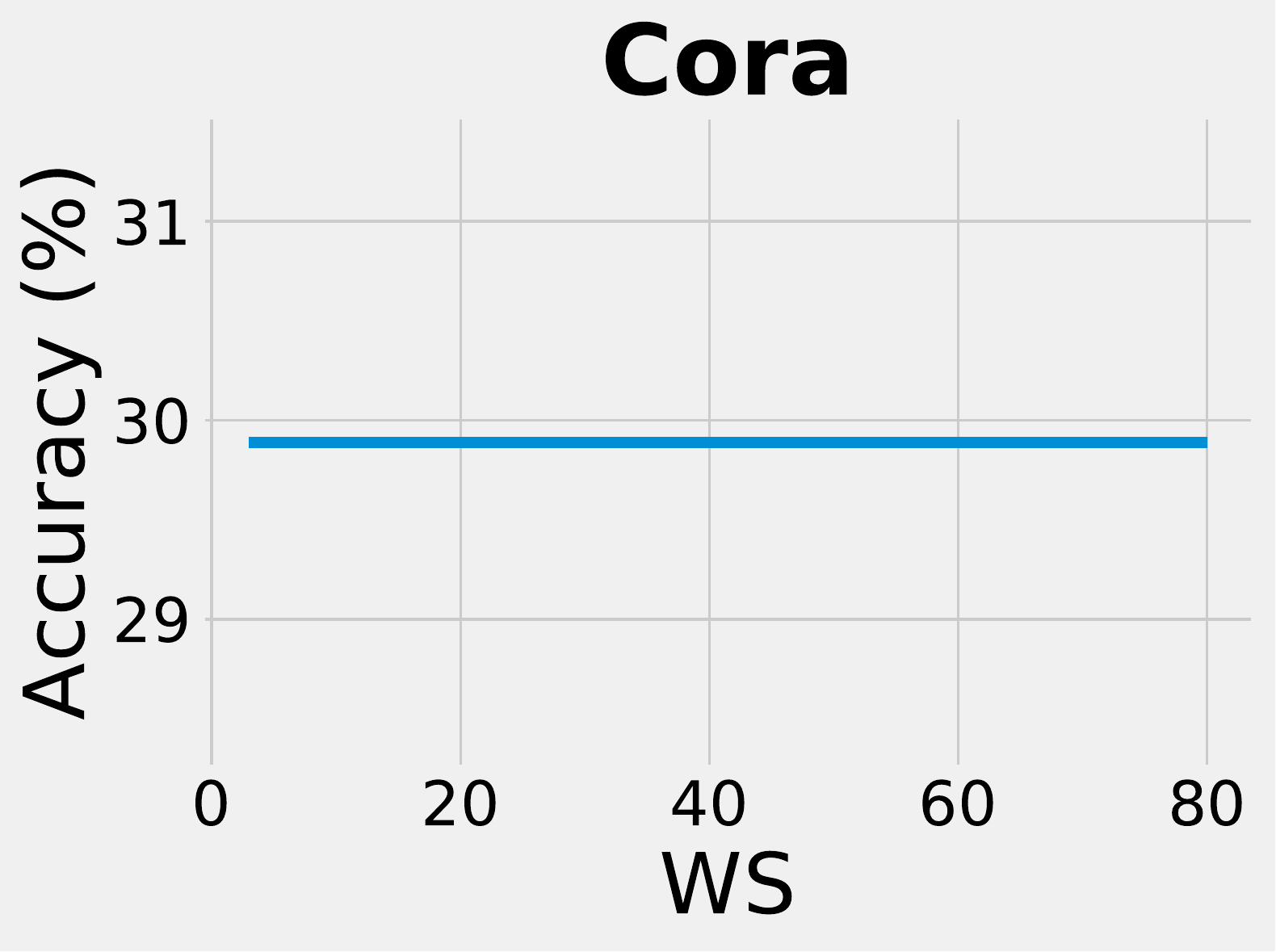}
  \end{subfigure}%
  \hfill
  \begin{subfigure}{0.14\textwidth}
  \centering 
    \includegraphics[width=\linewidth]{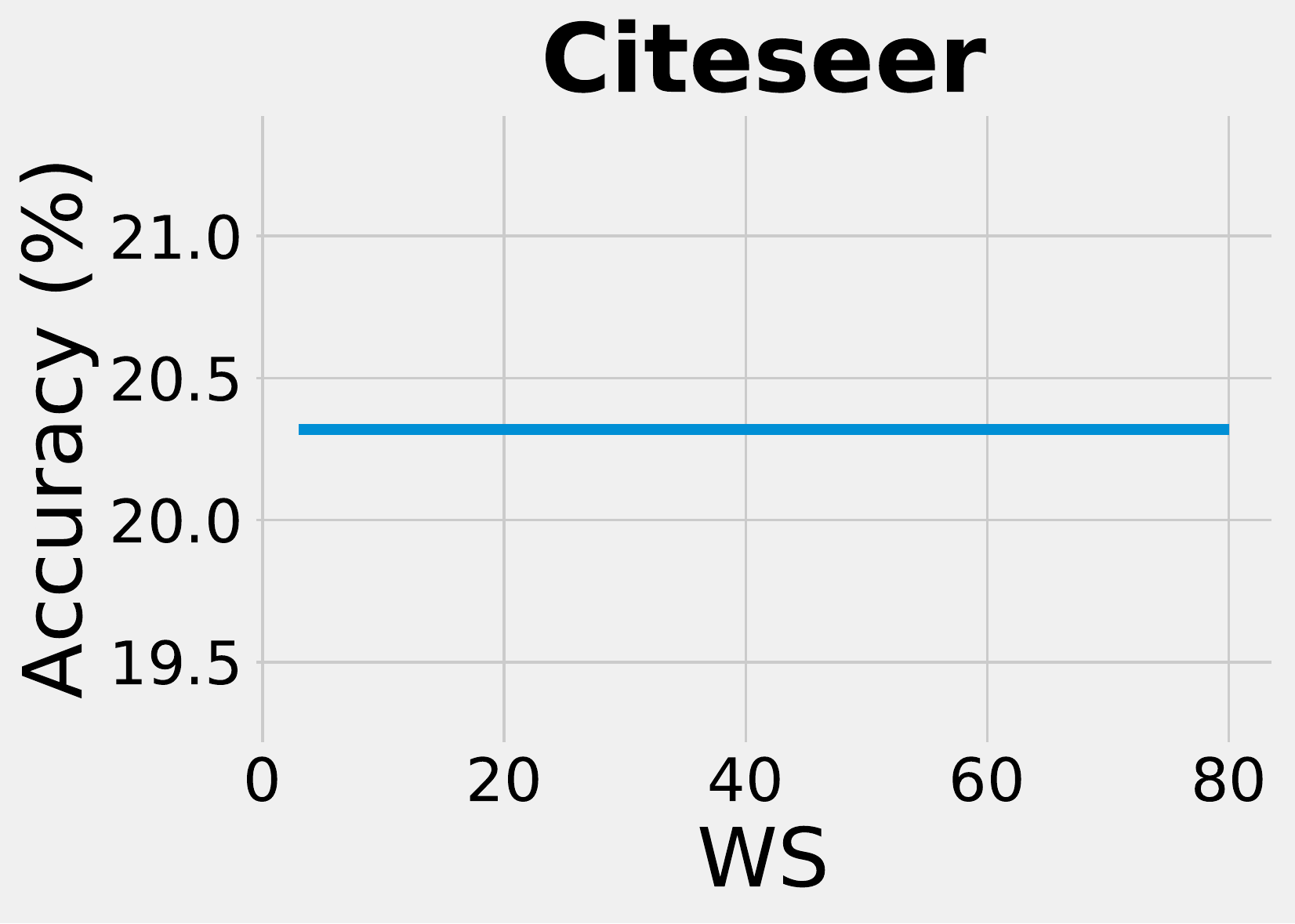}
  \end{subfigure}
  \hspace{-1.7mm}
  \begin{subfigure}{0.14\textwidth}
  \centering 
    \includegraphics[width=\linewidth]{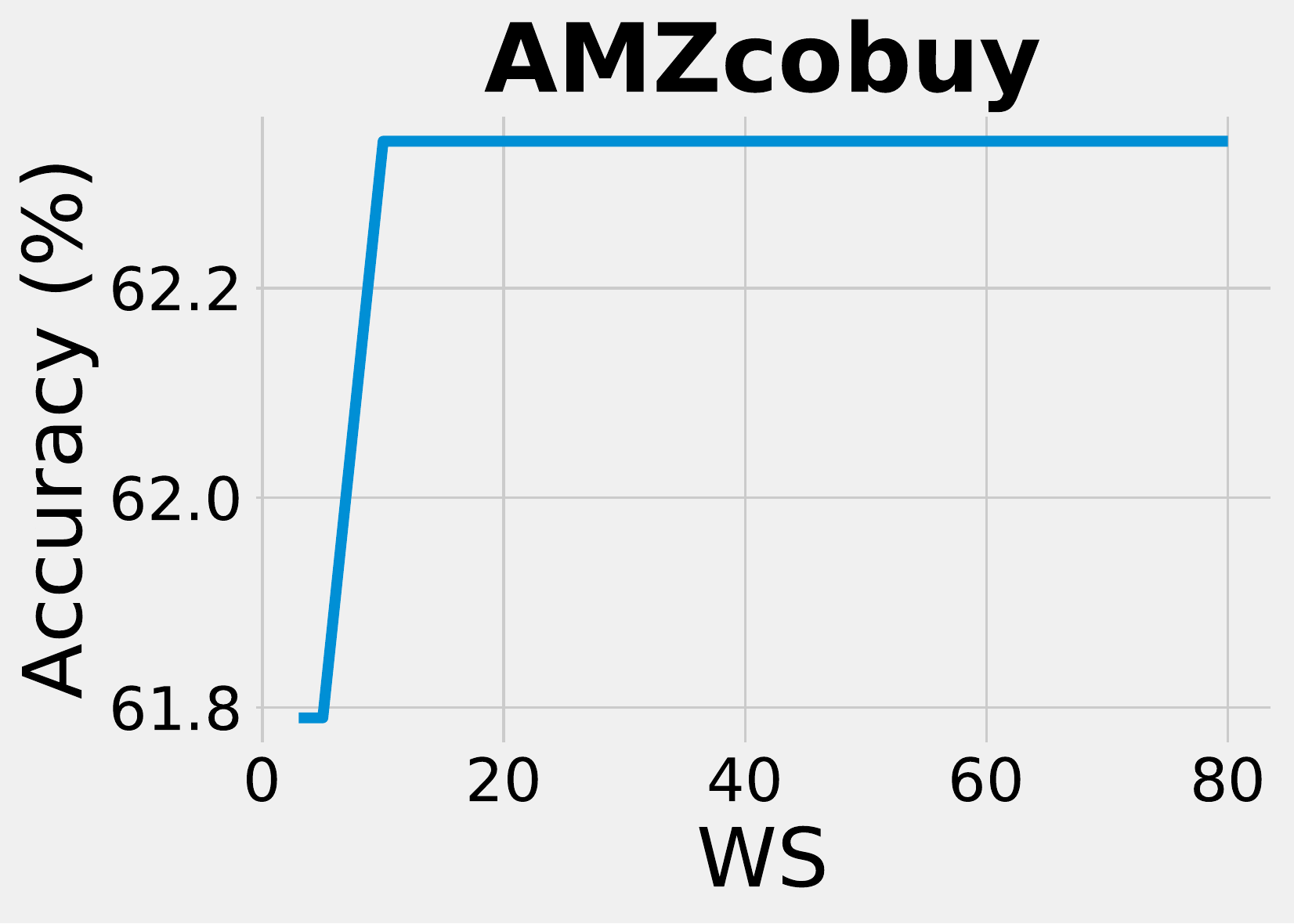}
  \end{subfigure}
  \hspace{-1.7mm}
  \begin{subfigure}{0.14\textwidth}
  \centering 
    \includegraphics[width=\linewidth]{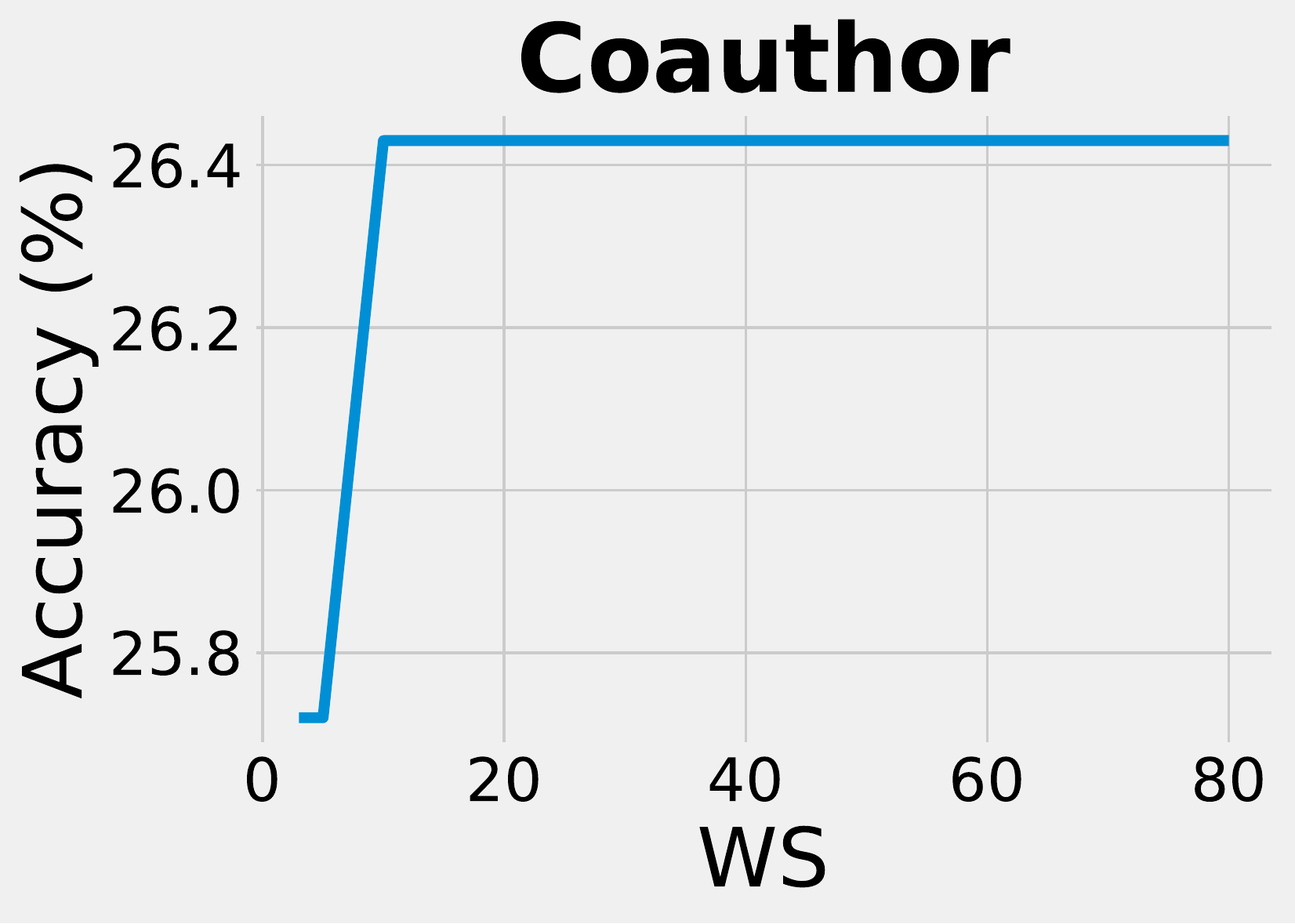}
  \end{subfigure}
  \hspace{-1.7mm}
  \begin{subfigure}{0.14\textwidth}
  \centering 
    \includegraphics[width=\linewidth]{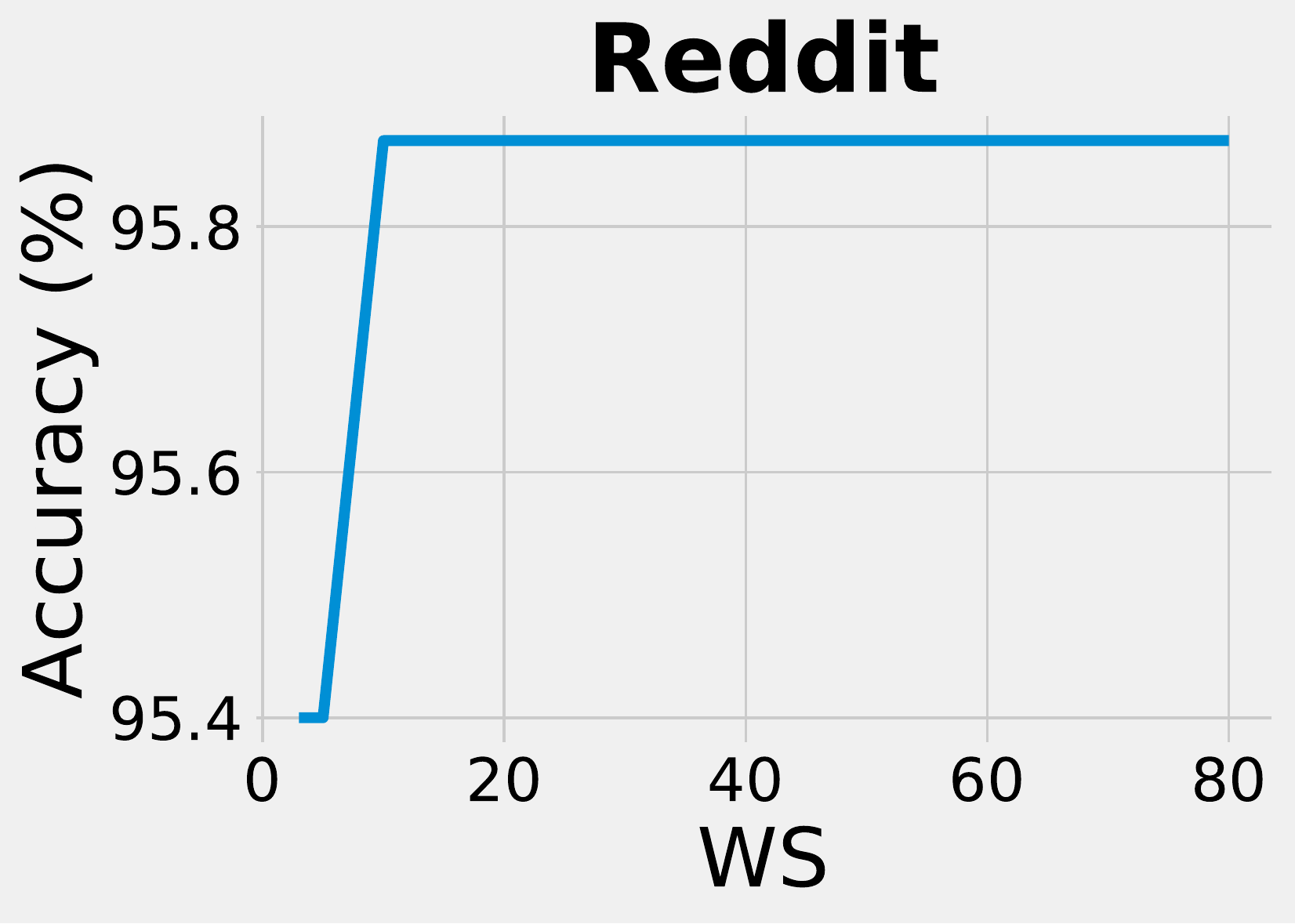}
  \end{subfigure}
\caption{Analysis of The Number of Warm-up Steps $WS$ for GraphLT Inference ($nr$=0.1)}
\label{fig:fig_ws}
\end{figure*}

\begin{table*}[t] 
\footnotesize
\centering
\setlength{\tabcolsep}{3pt}
\caption{Analysis of Average/Unit(Per 100 Nodes) Runtime for GraphLT Inference with Different $WS$ ($nr$=0.1)}
\label{table:runtime}
\begin{tabular}{cccccccc} 
\toprule
\textbf{Runtime (s)} & \textbf{KDD20(S1)} & \textbf{KDD20(S2)} & \textbf{Cora} & \textbf{Citeseer} & \textbf{AMZcobuy} & \textbf{Coauthor} & \textbf{Reddit} \\
\midrule
\textbf{Average} & 536.04 ($\pm$ 6.73) & 635.81 ($\pm$ 8.22) & 3.32 ($\pm$ 0.53) & 3.84 ($\pm$ 0.80) & 15.26 ($\pm$ 0.87) & 19.85 ($\pm$ 0.73) & 224.11 ($\pm$ 9.02) \\
\textbf{Unit} & 0.3289 & 0.3477 & 0.4089 & 0.3848 & 0.3699 & 0.3609 & 0.3206 \\
\bottomrule
    \end{tabular}
\end{table*}

\noindent
\textbf{Comparison of The Defense.}
Although all node classifiers benefit from label transition, GraphSAGE shows better performance before perturbation and also after label transition. 
For each scenario in Table \ref{table:exp1}, we highlight the best performance on each dataset with different colors. For example, GraphSAGE achieves the best performance across all three scenarios on KDD20(S1). So, we employ GraphSAGE as the node classifier of GraphLT in this section and further examine the defending performance between GraphLT and competing methods.
In Table \ref{table:performance}, we highlight the best defending performance (Red color) across seven datasets. The defending results indicate that GraphLT achieves superior defense against competing defenders on most benchmark datasets.
We observe that GRAND gains better performance on small graphs whereas works worse on larger graphs. One of the reasons for this is that GRAND assumes that the graph satisfies the homophily property. Larger graphs contain more nodes and a higher number of classes, which may make the network more difficult to satisfy homophily property leading to poor performance by GRAND. We also notice that RGCN gets worse performance on denser graphs. We argue that RGCN adopts a sampling in the hidden representation whereas this sampling may lose more information on denser graphs.
We also visualize the result of three scenarios for GraphLT on two large graphs, KDD20(S1) and Reddit, as examples in Figure \ref{fig:fig_cm}. The first three columns present the result of three scenarios. The last column shows the noisy labels with the noise ratio, $nr$ = 0.1. KDD20(S1) suffers from a serious class-imbalanced problem. Most uniform noises are distributed in three classes. The defending result on KDD20(S1) indicates that GraphLT can largely remedy the perturbation by label transition. Similarly, GraphLT shows satisfactory remedy on Reddit as well. We observe that one of the classes almost disappeared even if the graph is not perturbed. GraphLT can partially retrieve the prediction in this class. Overall, the visualization demonstrates that GraphLT can significantly improve the prediction on both datasets.

\noindent
\textbf{Analysis of Parameters.}
We analyze how the number of warm-up steps $WS$ affects accuracy. We conduct this analysis on the validation set. We notice that our model can achieve the best performance with only 100 epochs for inference. Thus, we fix the number of epochs for inference as 100 and examine the model with the different number of warm-up steps $WS$, where $WS \in [3, 80]$. In general, the change of $WS$ has a limited effect on accuracy. To enlarge the difference, we display the curve separately in Figure \ref{fig:fig_ws}. We observe that both larger and smaller $WS$ have a negative effect on accuracy. Larger $WS$ means insufficient inference, whereas smaller $WS$ implies inadequate epochs to build the dynamic label transition matrix $\phi$. This phenomenon shows more obviously on a large graph, such as KDD20(S1) or KDD20(S2). In this study, we select the $WS$ for KDD20(S1) and KDD20(S2) as 15 and 30, respectively. For the rest datasets, we fix the $WS$ as 20 instead.
Besides the $WS$, we analyze the runtime of GraphLT inference. The first row of Table \ref{table:runtime} presents the average runtime of GraphLT inference with different $WS$. It's expected that our model has longer runtime on a larger graph. We observe that the standard deviation for each dataset is stable, which indicates that the runtime stays stable no matter how the $WS$ changes. Besides, we also present the runtime per 100 validation nodes in the second row of Table \ref{table:runtime}. The result specifies that this unit runtime does not increase as the size of the graph grows. In other words, the speed of inference won't slow down on a larger graph.

\begin{figure}[t] 
  \centering
  \includegraphics[width=\linewidth]{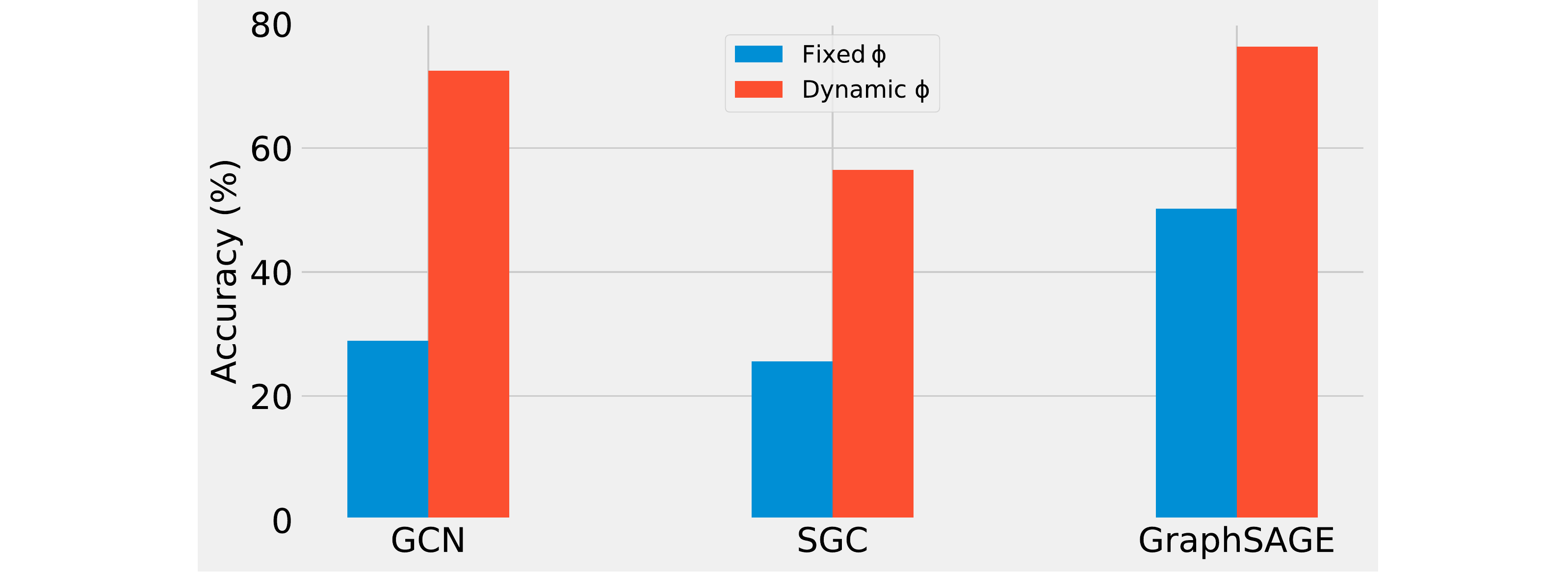}
  \caption{Ablation Study of GraphLT over Three Node Classifiers on KDD20(S1) (\{Fixed $\phi$, Dynamic $\phi$\} denote whether we dynamically update the transition matrix $\phi$ in each iteration.)}
\label{fig:fig_bar}
\end{figure}

\noindent
\textbf{Ablation Study.}
We conduct an ablation study over three node classifiers on KDD20(S1) to illustrate how the label transition matrix $\phi$ affects the defending performance of GraphLT. We follow the same procedure as our previous defense experiments.
According to Figure \ref{fig:fig_bar}, we observe that the accuracy has no change without dynamically updating the transition matrix $\phi$. This study reveals that dynamic transition matrix $\phi$ is a crucial component, which can significantly improve the performance of GraphLT.

\noindent
\textbf{Limitation and Future Directions.}
The inference of GraphLT depends on the warm-up label transition matrix $\phi'$, which is built with the categorical distribution on the train graph. Such dependence indicates that GraphLT can only handle the perturbations on the test graph at this point.
In the future, GraphLT can be improved to defend the poisoning perturbations by iteratively updating the noisy labels. Besides, our work could be easily extended to detect the anomaly perturbators on dynamic graphs.

\section{Related Work}
\label{sec:rewk}
Related works about the defense of GCNs are already discussed in {\bf Introduction}. Most existing studies about learning with noisy labels achieve great accomplishments on both image and graph domain \cite{sukhbaatar2014training, misra2016seeing, yao2019safeguarded, reed2014training, nt2019learning, zhang2020adversarial, dai2021nrgnn, xia2020towards}. In this work, we propose a Bayesian label transition model to improve the performance of the node classifier on graph data without utilizing adversarial samples or identifying perturbators.

\section{Conclusion}
\label{sec:con}
In this paper, we propose a new Bayesian label transition model, namely GraphLT, to improve the performance of OSN classifier by approximating the inferred labels to the latent labels based on dynamic conditional label transition, which follows Dirichlet distribution.
GraphLT provides two main advantages. For one thing, GraphLT can significantly reverse the performance's decline caused by training a node classifier with noisy labels. For another, GraphLT can repair the prediction of the node classifier under a perturbed environment without identifying perturbators. Extensive experiments demonstrate these two advantages over seven benchmark datasets.

\bibliographystyle{siamplain}
\bibliography{reference}

\appendix

\section{Reproducibility}
\subsection{Hardware and Software.}
All experiments are conducted on Ubuntu 18.04.5 LTS with Intel(R) Xeon(R) Gold 6258R 2.70 GHz CPU and NVIDIA Tesla V100 PCIe 16GB GPU.

\subsection{Hyper-parameters of Our Model.}
Our proposed model, GraphLT, can be applied on top of GCNs. The model architecture and hyper-parameters of GCNs are described in Table \ref{tab:gcn_archit} and Table \ref{tab:hp}, respectively. In the training phase, the number of training epochs for all node classifiers is kept the same for each graph. The hyper-parameter of GraphLT, $\alpha$, is fixed as 1.0.

\begin{table}[t]
\centering
\caption{Hyper-parameters of GCNs (\#Hidden denotes the number of neurons in each hidden layer of GCNs)}
\begin{tabular}{cc}
  \toprule
    \textbf{Hyper-parameters} & \textbf{Values} \\
    \midrule
    \textbf{\#Layers} & 2 \\
    \textbf{\#Hidden} & 200 \\
    \textbf{Optimizer} & Adam \\
    \textbf{Learning Rate} & $1 \times 10^{-3}$ \\
  \bottomrule
\end{tabular}
\label{tab:hp}
\end{table}

\begin{table}[t]
\footnotesize
\centering
\setlength{\tabcolsep}{2.8pt}
\caption{Model Architecture of GCNs}
\begin{tabular}{ccccc}
 \toprule
  \textbf{Model} & \textbf{Aggregator} & \textbf{\#Hops} & \textbf{Activation} & \textbf{Dropout} \\
  \midrule
    $\mathbf{GCN}$ & $\times$ & $\times$ & ReLU & 0.0 \\
    $\mathbf{SGC}$ & $\times$ & 2 & $\times$ & 0.0 \\
    $\mathbf{GraphSAGE}$ & mean & $\times$ & ReLU & 0.0 \\
 \bottomrule
\end{tabular}
\label{tab:gcn_archit}
\end{table} 

\subsection{Hyper-parameters of Competing Methods.}
\begin{figure}[h]
  \centering
  \includegraphics[width=\linewidth]{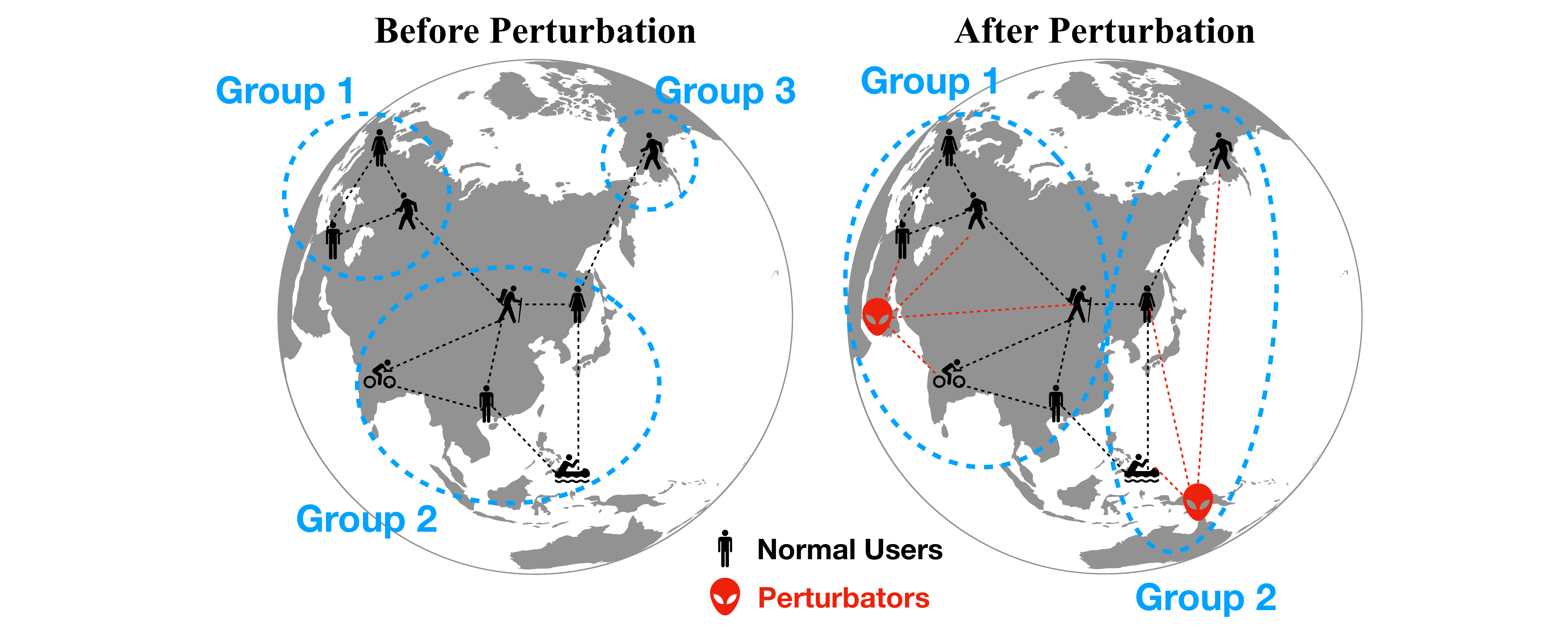}
  \caption{An Example of Non-malicious Perturbations. (The left-hand side presents the OSN before perturbation (unperturbed environment). The node classifier can classify normal users into the correct group. The right-hand side shows the OSN after perturbation. Perturbators randomly connect with many other users and disturb the accuracy of the node classifier, causing incorrect classification. The red dashed line denotes the connection between the normal user and the perturbator.)}
\label{fig:perturbation}
\end{figure}

We examine the defending performance between our model and competing methods against the non-malicious perturbations, which is described in Figure \ref{fig:perturbation} as an example. For reproducibility purposes, we maintain the same denotation for each competing method as the corresponding original paper and present the hyper-parameters below. The competing models are trained by Adam optimizer with 200 epochs.
\begin{description}
\item[$\bullet$\ RGCN~\cite{zhu2019robust}] adopts Gaussian distributions as the hidden representations of nodes to mitigate the negative effects of adversarial attacks. We set up $\gamma$ as 1, $\beta_1$ and $\beta_2$ as $5 \times 10^{-4}$ on all datasets. The number of hidden units (\#hidden) is 32. The dropout rate is 0.6. The learning rate is 0.01.
\item[$\bullet$\ GRAND~\cite{feng2020graph}] proposes random propagation and consistency regularization strategies to address the issues about over-smoothing and non-robustness of GCNs. We follow the same procedure to tune the hyper-parameters and present them in Table~\ref{tab:grand_para}.
\item[$\bullet$\ ProGNN~\cite{jin2020graph}] jointly learns the structural graph properties and iteratively reconstructs the clean graph to reduce the effects of adversarial structure. We select $\alpha$, $\beta$, $\gamma$, and $\lambda$ as $5 \times 10^{-4}$, 1.5, 1.0, and $1 \times 10^{-3}$, respectively. \#hidden is 16. The dropout rate is 0.5. The learning rate is 0.01. Weight decay is $5 \times 10^{-4}$.
\item[$\bullet$\ GNNGUARD~\cite{zhang2020gnnguard}] employs neighbor importance estimation and the layer-wise graph memory to quantify the message passing between nodes to defend the GCNs from adversarial attacks. We follow the same setting as \cite{zhang2020gnnguard}. \#hidden is 16. The dropout rate is 0.5. The learning rate is 0.01.
\end{description}

\begin{table}[h]
\footnotesize
\centering
\setlength{\tabcolsep}{3pt}
  \caption{Hyper-parameters of GRAND in This Paper for Small (Cora, Citeseer), Medium (AMZcobuy, Coauthor), and Large graphs (KDD20(S1), KDD20(S2), and Reddit)}
  \label{tab:grand_para}
  \begin{tabular}{cccc}
    \toprule
    \textbf{Hyperparameters} & \textbf{Small} & \textbf{Medium} & \textbf{Large} \\
    \midrule
    DropNode probability & 0.5 & 0.5 & 0.5 \\
    Propagation step & 8 & 5 & 5 \\
    Data augmentation times & 4 & 4 & 3\\
    CR loss coefficient & 1.0 & 1.0 & 0.9 \\
    Sharpening temperature & 0.5 & 0.2 & 0.4 \\ 
    Learning rate & 0.01 & 0.2 & 0.2 \\
    Early stopping patience & 200 & 100 & 100 \\
    Hidden layer size & 32 & 32 & 32 \\
    L2 weight decay rate & $5 \times 10^{-4}$ & $5 \times 10^{-4}$ & $5 \times 10^{-4}$ \\
    Dropout rate in input layer & 0.5 & 0.6 & 0.6 \\
    Dropout rate in hidden layer & 0.5 & 0.8 & 0.5 \\
  \bottomrule
\end{tabular}
\end{table} 

\end{document}